\documentclass{article}

\usepackage{microtype}
\usepackage{graphicx}
\usepackage{subcaption}
\usepackage{booktabs}
\usepackage{placeins}

\usepackage{hyperref}

\usepackage[preprint]{icml2026}

\usepackage[utf8]{inputenc}
\usepackage[T1]{fontenc}
\usepackage{url}
\usepackage{amsmath}
\usepackage{amssymb}
\usepackage{amsfonts}
\usepackage{mathtools}
\usepackage{amsthm}
\usepackage{nicefrac}
\usepackage{xcolor}

\graphicspath{{./}}

\icmltitlerunning{Self-Play RL for Big 2}

\begin{document}

\twocolumn[
  \icmltitle{Self-Play Reinforcement Learning under Imperfect Information in Big 2}

  \begin{icmlauthorlist}
    \icmlauthor{Aalok Patwa}{upenn}
  \end{icmlauthorlist}

  \icmlaffiliation{upenn}{University of Pennsylvania}
  \icmlcorrespondingauthor{Aalok Patwa}{aalok.patwa7@gmail.com}

  \icmlkeywords{Reinforcement Learning, Self-Play, Imperfect-Information Games, Card Games}

  \vskip 0.3in
]

\printAffiliationsAndNotice{}

\begin{abstract}
Imperfect-information multiplayer games test whether agents can act under hidden information, sparse rewards, and non-stationary opponents. We study these challenges in Big 2, a four-player imperfect-information card game. We develop a self-play RL framework for Big 2 that enables controlled comparisons between policy-gradient and value-approximating agents. Under a common environment, input representation, training budget, and evaluation protocol, PPO outperforms Monte Carlo Q approximation, SARSA, and Q-learning against random, greedy, and heuristic Big 2 opponents. We further find that moderate entropy regularization improves PPO by preventing the policy from becoming overly deterministic, and that current-policy self-play provides a stronger finite-budget curriculum than checkpoint self-play or fixed-opponent training. Together, these results show that Big 2 is a useful controlled setting for studying deep RL under imperfect information, multiplayer interaction, delayed rewards, and variable action sets.

\end{abstract}

\section{Introduction}

Games are a useful testbed for reinforcement learning (RL) because they provide precise rules, rewards, and evaluation protocols. In perfect-information games, self-play RL and search have produced numerous successes, from AlphaGo to AlphaZero and MuZero \citep{silver2016alphago,silver2018alphazero,schrittwieser2020muzero}. Imperfect-information games are harder: agents must act from partial observations, infer hidden state from public behavior, and learn under non-stationary opponent distributions induced by self-play. Progress in poker, Hanabi, Stratego, and general game-playing systems has shown the power of combining learning with search, regret minimization, or game-theoretic reasoning \citep{heinrich2016nfsp,moravcik2017deepstack,brown2018libratus,brown2019deepcfr,brown2019pluribus,brown2020rebel,bard2020hanabi,perolat2022stratego,schmid2023studentgames}. Recent work on action abstraction and policy-gradient theory highlights the need for better understanding of learning dynamics in imperfect-information games \citep{li2024rlcfr,liu2025policygradient}.

Multiplayer card games provide a challenge for game-theoretic learning algorithms, as they involve hidden information, sparse terminal rewards, and action spaces that change sharply between turns of play. For example, the game DouDizhu features three-player competition and cooperation and a large variable action space \citep{zha2021douzero}, Mahjong requires reasoning about hidden information across four players \citep{li2020suphx}, and Pluribus showed that moving beyond heads-up poker introduces qualitatively new strategic issues \citep{brown2019pluribus}.

We study Big 2, a four-player card-shedding game. Each player observes only their own hand and the public play history, while the other three hands must be inferred from actions, passes, and remaining card counts. The game's legal actions are hand-specific combinations such as singles, pairs, triples, straights, flushes, full houses, four-of-a-kind hands, straight flushes, and passes. Prior Big 2 work has demonstrated the difficulty of mastering the game due to multiplayer dynamics, large state and action spaces, and short-term versus long-term strategic tradeoffs \citep{chen2022big2ai,luo2024odmc,chen2025big2mdp}. Big 2 is particularly challenging because playing a strong short-term action may greatly reduce a player's future options or allow an opponent to take control of the game. Therefore, the game tests whether an agent can choose the long-term strategic action over the locally optimal action.

Prior Big 2 agents have used self-play PPO, Monte Carlo tree search-based opponent prediction, Monte Carlo training with opponent modeling and action filtering, and MDP-style decompositions of scoring, risk, prediction, and control \citep{charlesworth2018selfplay,chen2022big2ai,luo2024odmc,chen2025big2mdp}. Our goal is complementary: we study compute-efficient deep RL methods in the full four-player environment, avoiding engineered opponent models, tree search, and heuristic action pruning beyond legal-action filtering.

Despite recent progress, there has not yet been a controlled study to investigate whether policy-gradient or value-based objectives learn more effectively in Big 2 under the same interface and limited training budget, and how training design choices affect stability and final performance. We compare PPO, Monte Carlo Q approximation, SARSA, and target-network Q-learning under a common environment, state and action representation, architecture, training budget, and evaluation protocol. This limited-compute setting lets us study sample and compute efficiency rather than performance gains from scale alone. We find that PPO performs best among the methods tested, and we analyze two training factors that substantially influence its performance: entropy regularization, which affects policy stochasticity, and opponent curriculum, which changes the learning signal. Together, these contributions provide the first controlled empirical study of RL objectives and training design choices for Big 2, as well as an accessible baseline for future work on search, abstraction, opponent modeling, and larger training budgets.

\section{Game Formulation}
We model Big 2 as a finite-horizon, turn-based, imperfect-information game with $N=4$ players and a standard 52-card deck. Card values are ranked in the order of 3, 4, 5, 6, 7, 8, 9, 10, J, Q, K, A, 2, and suits break ties in the order diamonds $<$ clubs $<$ hearts $<$ spades. Each player receives 13 private cards, the player holding $3\diamondsuit$ opens, and players act clockwise until one player empties their hand and wins. A trick is the current combination that other players must beat or pass on. Legal non-pass tricks are singles, pairs, triples, and five-card hands; five-card hands are ordered as straight $<$ flush $<$ full house $<$ four-of-a-kind $<$ straight flush. A response to a single, pair, or triple must be a trick of the same category with higher value, while a response to a five-card hand must either beat it within the same category or use a higher five-card category. If all other players pass after a non-pass play, the trick is cleared and the last player to play regains control, meaning they may lead another round of play with any legal non-pass combination. The goal is for a player to be the first to discard all of their cards.

\section{Methods}
\subsection{Game Environment}
\label{sec:candidate_scoring}
We developed a simulator whose observation state $o_i$ accurately reflects the information available to each player during the game. At each decision point, the acting player observes their own private hand and the public game history, including the active trick, previously played cards, remaining card counts of each opponent, and the current pass count. The simulator also returns the current legal candidate action set $\mathcal{A}(o_i)$ by enumerating legal combinations in the acting player's hand and filtering to those that would be valid given the active trick. This avoids invalid-action exploration and makes policy-gradient and value-based methods directly comparable despite the variable action set. Each candidate action is represented as a feature vector containing rank-and-suit indicator features, a bit for whether the action is a pass, the trick type, and trick rank features. Further details are provided in Appendix~\ref{app:game_details}.

\subsection{Neural Architecture}
Our architecture separately encodes the information state and the legal candidate actions, then scores each state-action pair. The acting player's hand is represented as card IDs, embedded with a shared card embedding table, passed through self-attention over the held cards, and pooled into a hand summary. Public card sets such as the current trick, seen cards, and opponents' played cards are embedded with the same table, concatenated with opponent card counts and the pass count, and projected into a state embedding. Legal actions are encoded from their 80-dimensional combination features and scored against the state embedding by a dot product, so the policy and Q-network rank only the actions that are legal at the current decision point. This differs notably from previous Big 2 PPO architectures, which feed a hand-engineered 412-bit state vector into fully connected layers and predict over a fixed 1695-action head that is masked to legal moves \citep{charlesworth2018selfplay}; our model represents cards directly, allows held cards to interact before pooling, and avoids predicting scores for illegal actions.

We implement a policy network that uses state-action scores as logits to calculate action probabilities, and also uses a separate MLP as a value head. We also implement a Q-network that directly uses state-action scores as Q values.

\begin{figure*}[!t]
    \centering
    \includegraphics[width=0.9\textwidth,height=0.55\textheight,keepaspectratio]{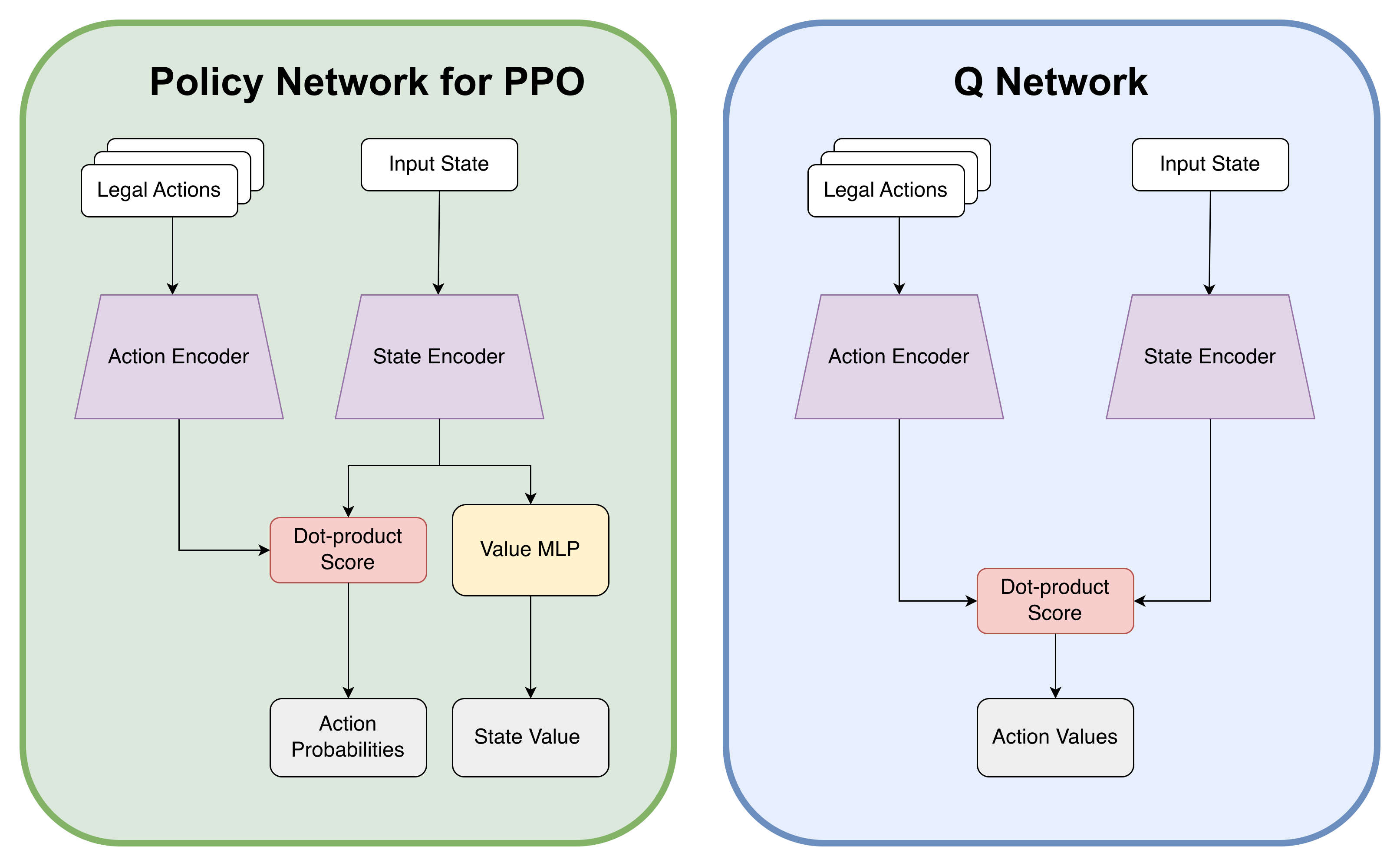}
    \caption{Neural architectures for different learning algorithms. The PPO policy and Q-network share the same card-aware state encoder, action encoder, and dot product scorer. The PPO policy network includes a value head.}
    \label{fig:model_architecture}
\end{figure*}

\subsection{Learning Algorithms}
We use PPO as our policy gradient baseline to train the policy network. We compare PPO with three value-based algorithms that collect trajectories using $\epsilon$-greedy action selection over legal actions and are trained using mean squared error on the predicted value of the selected action. Our reward signal is based on the Big 2 game score: the winner of a game receives a game score equal to the sum of the remaining cards in the losers' hands, and each loser receives a score equal to the negative of their remaining card count. For training the value-based algorithms only, the reward is divided by 13 as explained in Appendix~\ref{app:training_details}. 

The Monte Carlo Q variant uses the full discounted return from each model-controlled trajectory,
\[
y_t = \sum_{k=t}^{T} \gamma^{k-t} r_k.
\]
The SARSA variant uses the one-step on-policy target,
\[
y_t = r_t + \gamma Q_{\mathrm{target}}(o_{t+1}, a_{t+1}),
\]
where $a_{t+1}$ is the next action actually selected by the behavior policy at the next model-controlled decision point. The Q-learning variant uses the corresponding max target,
\[
y_t = r_t + \gamma \max_{a \in \mathcal{A}(o_{t+1})} Q_{\mathrm{target}}(o_{t+1}, a).
\]
For the SARSA and Q-learning variants, a delayed target network is periodically synchronized with the online Q-network.

All three value-based agents behave greedily during evaluation.

\section{Experimental Setup}
\paragraph{Training configuration.}
To study learning dynamics, we train each agent in a limited-compute setting: 5,000 batches at 64 episodes per batch. Across all algorithms evaluated, these 5{,}000 batches took between 7 hours and 13 hours to train on a single 6-core Intel i7 laptop. 

For PPO, we use 4 PPO epochs per update, clip $\epsilon=0.2$, learning rate $3 \times 10^{-5}$, $\gamma=0.99$, and $\lambda=0.95$. We use learning rate warmup and cosine learning rate decay. Additional implementation details for PPO and value-based training are in Appendix~\ref{app:training_details}.

During current-policy self-play, the current policy controls all four seats, and training examples are collected from every model-controlled decision point across seats. When training against a fixed opponent, the policy occupies one randomly chosen seat and the fixed opponent controls the other three.
\paragraph{Evaluation protocol.}
To provide a consistent evaluation baseline, we implement three heuristic opponents of varying difficulty. The first is a "Random" baseline, which simply chooses uniformly from the legal action set. The second is a "Greedy" baseline, described in Algorithm~\ref{alg:greedy_strategy}, which plays the weakest legal non-pass combination available. The third is a "Smart" baseline, described in Algorithm~\ref{alg:smart_strategy}, which is a stronger hand-aware rule-based policy. It scores each legal non-pass action using lightweight strategic features, including immediate wins, number of cards shed, and whether it leaves low orphan singles. It only passes in narrow situations, such as to avoid expensive early use of 2s and conserve valuable five-card combinations.

At evaluation time, we roll out 1{,}000 four-player games where one seat is held by the agent being evaluated, and the other three seats are held by players of a certain opponent class. The evaluated agent's seat is randomized across games, and all reported metrics are averaged over these seat-randomized deals. We track win rate and the average game score (reward) against each opponent class. Therefore, we consider an agent successful against an opponent pool when its win rate exceeds 25\% and its average score is positive. Unless otherwise noted, each reported evaluation result is from a single training and evaluation seed. Because our result tables do not report uncertainty across independent seeds, we interpret small differences cautiously.

\section{Results}
\subsection{PPO outperforms value-based methods under self-play}
\label{sec:results_learning_curves}
\begin{figure*}[!t]
    \centering
    \includegraphics[width=0.9\textwidth,height=0.68\textheight,keepaspectratio]{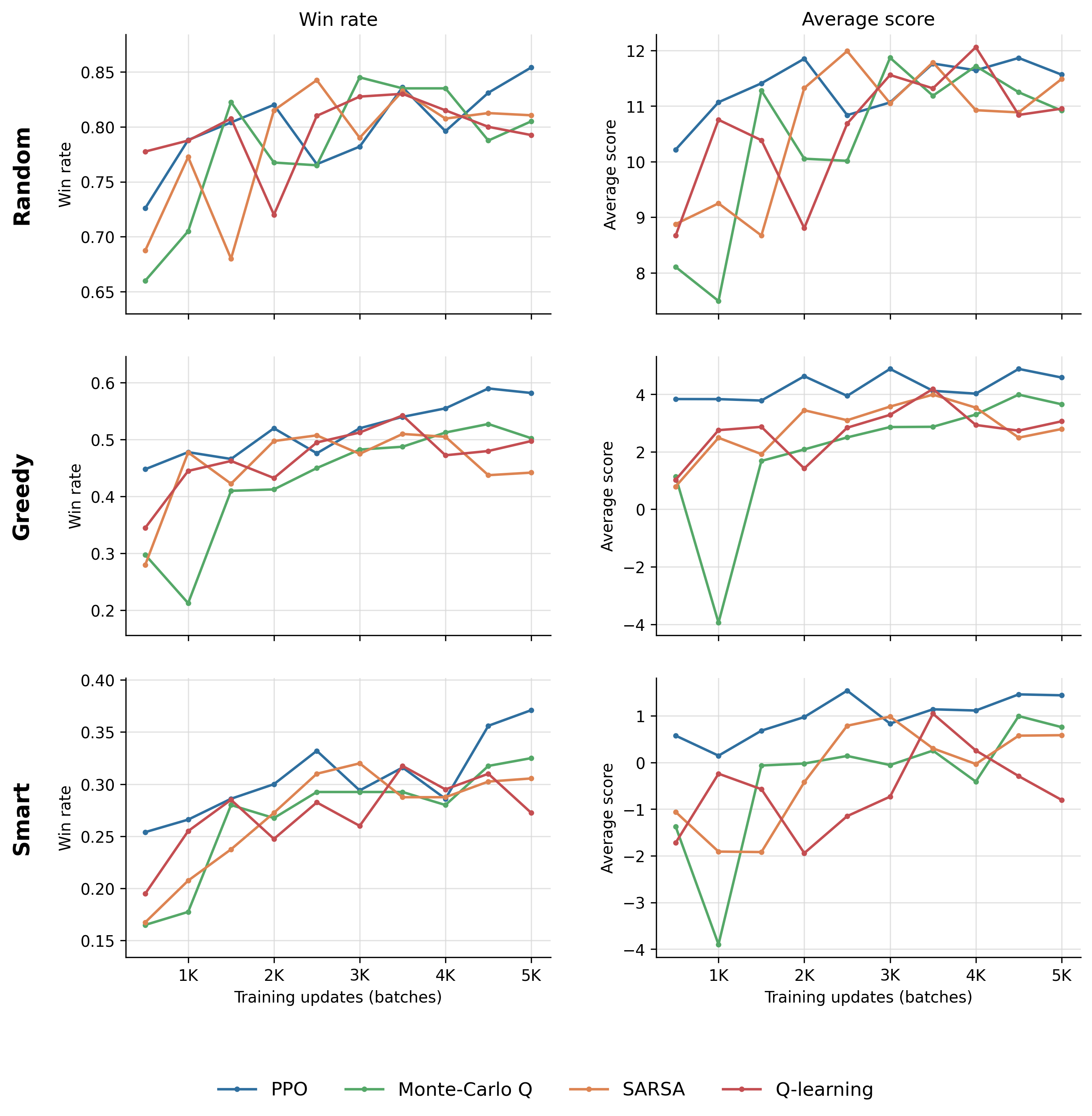}
    \caption{Learning curves for PPO, Monte Carlo Q, SARSA, and Q-learning in four-player Big 2. Each checkpoint is evaluated against fixed random, greedy, and smart heuristic opponent pools. We report win rate and average score for the model-controlled seat.}
    \label{fig:training_performance_overview}
\end{figure*}

Figure~\ref{fig:training_performance_overview} compares the training dynamics of the policy-gradient and value-based agents under a shared simulator, representation, and evaluation protocol. PPO is the strongest and most consistent method over the training budget analyzed, and Table~\ref{tab:final_checkpoint_performance} shows that it has the best final-checkpoint win rate and average score against all three opponent classes. It improves rapidly early in training and remains competitive across all three opponent pools, with especially clear gains against the greedy and smart heuristic opponents. This finding is consistent with prior Big 2 work showing that self-play PPO can learn robust strategies in the game \citep{charlesworth2018selfplay}, but our comparison extends that observation by evaluating PPO alongside multiple value-based alternatives under the same legal-candidate scoring interface and compute budget.

The value-based methods learn useful policies, but they do not match PPO's overall performance within the same training horizon. Among these methods, Monte Carlo Q is the strongest final-checkpoint value-based baseline, outperforming SARSA and Q-learning against the greedy and smart opponents. The gap between PPO and the value-based methods contrasts with DouZero, where Monte Carlo value approximation was highly effective for DouDizhu self-play \citep{zha2021douzero}, and with Big 2 DMC variants that achieve strong performance through longer training, opponent modeling, and action-set filtering \citep{luo2024odmc}. One plausible explanation is that Big 2's large, four-player, hidden-information state space makes value estimation slow to stabilize: strategically important states are visited rarely, and terminal returns must assign credit across long sequences of combinatorial actions. Under this interpretation, PPO's clipped policy-gradient objective and value baseline provide a more sample-efficient learning signal for the training budgets we study, while Monte Carlo Q may require longer training or additional structure such as opponent modeling and action pruning to close the gap.

The results also suggest that self-play is not merely overfitting to a single opponent distribution. Although training uses self-play rather than direct supervised imitation of the evaluation opponents, performance improves against Random, Greedy, and Smart heuristics. This cross-opponent improvement suggests that the agents, and PPO in particular, learn transferable Big 2 strategies.

\begin{table}[!tbp]
\centering
\caption{Win rates and average score across algorithms.}
\label{tab:final_checkpoint_performance}
\begingroup
\scriptsize
\setlength{\tabcolsep}{2.5pt}
\renewcommand{\arraystretch}{0.9}
\begin{tabular}{@{}lccc@{}}
\toprule
Method & Random & Greedy & Smart \\
\midrule
PPO & \textbf{85.4\% (11.56)} & \textbf{58.2\% (4.58)} & \textbf{37.1\% (1.44)} \\
Monte Carlo Q & 80.5\% (10.92) & 50.2\% (3.65) & 32.5\% (0.76) \\
SARSA & 81.0\% (11.48) & 44.2\% (2.79) & 30.6\% (0.58) \\
Q-learning & 79.2\% (10.95) & 49.8\% (3.06) & 27.3\% (-0.80) \\
\bottomrule
\end{tabular}%
\endgroup
\end{table}

\subsection{Moderate entropy regularization improves PPO}
The PPO results in Section~\ref{sec:results_learning_curves} were obtained from a run with no entropy regularization. After inspecting model outputs, we found that the average policy entropy decreased steadily over training, as shown in Appendix Figure~\ref{fig:ppo_entropy_ablation}. This was confirmed by examining evaluation rollouts, which showed that the model often sampled its top action with 90+\% probability, including in ambiguous but strategically important decision points, such as the first trick in the game. This behavior suggested that the lack of entropy regularization may have made the policy too deterministic. In an imperfect-information game such as Big 2, stochastic policies may perform better as they operate with uncertainty given hidden information and avoid becoming predictable.

We therefore ablate the effect of explicitly encouraging stochasticity in PPO. For these runs, we add the standard entropy term to the PPO minimization objective,
\[
L_{\mathrm{PPO}} = L_{\mathrm{policy}} + c_v L_{\mathrm{value}} - \beta_{\mathrm{ent}} \mathbb{E}_{o}\left[H\left(\pi(\cdot \mid o)\right)\right],
\]
where $\beta_{\mathrm{ent}}$ controls the strength of the entropy incentive. Appendix Figure~\ref{fig:ppo_entropy_ablation} shows that increasing $\beta_{\mathrm{ent}}$ does in fact make the trained policy maintain stochasticity throughout training.

Table~\ref{tab:ppo_entropy_ablation_performance} reports the final performance of PPO agents trained using different entropy incentives. $\beta_{\mathrm{ent}} = 0.05$ achieves the best performance, suggesting that moderate entropy regularization improves performance, but only to an extent; too much entropy trades off with learning better policies and potentially leads the model to take suboptimal actions.

\begin{table}[!t]
\centering
\caption{Results of PPO entropy ablation.}
\label{tab:ppo_entropy_ablation_performance}
\resizebox{\columnwidth}{!}{%
\begin{tabular}{lccc}
\toprule
Entropy beta & Random & Greedy & Smart \\
\midrule
$0.00$ & 85.4\% (11.56) & 58.2\% (4.58) & 37.1\% (1.44) \\
$0.05$ & \textbf{90.1\% (13.10)} & \textbf{64.8\% (5.40)} & \textbf{43.5\% (2.00)} \\
$0.10$ & 87.0\% (12.63) & 60.3\% (5.10) & 39.7\% (1.70) \\
\bottomrule
\end{tabular}%
}
\end{table}

\FloatBarrier

\subsection{Current-policy self-play outperforms alternative curricula}
We next ablate the opponent distribution used during training. The default setting trains against the current policy, which exposes the learner to an opponent distribution that changes as the agent improves. We compare this setting to checkpoint self-play, where opponents are sampled from earlier saved policies in the same training run, and to a fixed-opponent curriculum in which the learning agent plays only against the deterministic Smart strategy. Results use $\beta_{\mathrm{ent}} = 0.05$ from above.

Table~\ref{tab:opponent_curriculum_ablation} shows that current-policy self-play performs best for both PPO and Monte Carlo Q under the training budgets we study. This result is somewhat counterintuitive for evaluation against the Smart opponent. In the limit of sufficient exploration, data, representation capacity, and optimization, a Smart-only curriculum could in principle learn a best response to Smart.

However, we hypothesize that training only against Smart produces a narrower distribution of states and legal action sets because the deterministic opponent repeatedly drives games through the parts of the game tree favored by its heuristic. This can reduce exploration and action-value coverage, especially for Monte Carlo Q, where terminal returns provide high-variance labels only for the actions actually sampled. Current-policy self-play, by contrast, acts as a moving curriculum: the agent sees weak opponents early and increasingly stronger opponents as its own policy improves. This keeps the opponent distribution close to the learner's current skill level, so rollouts tend to expose mistakes that are still relevant to the current policy.

Checkpoint self-play also adds opponent diversity, but it weakens this adaptive pressure. Older checkpoints can represent behaviors that the current policy has already learned to beat, so part of the training budget is spent collecting gradients against stale mistakes rather than against the learner's present strategic weaknesses. The checkpoint pool therefore trades the sharper learning signal from current-policy opponents for broader but less targeted opponent coverage. Under longer training this diversity may improve robustness, but in our limited-training-budget setting the diluted learning signal is slightly worse than training directly against the current policy.

\begin{table}[!t]
\centering
\caption{Opponent-curriculum ablation. Entries show win rate and average score.}
\label{tab:opponent_curriculum_ablation}
\resizebox{\columnwidth}{!}{%
\begin{tabular}{lccc}
\toprule
Training curriculum & Random & Greedy & Smart \\
\midrule
\multicolumn{4}{l}{\textit{PPO}} \\
Current self-play & \textbf{90.1\% (13.10)} & \textbf{64.8\% (5.40)} & \textbf{43.5\% (2.00)} \\
Checkpoint self-play & 88.7\% (11.74) & 61.6\% (4.78) & 40.6\% (1.44) \\
Smart-only & 83.9\% (11.58) & 55.4\% (4.55) & 37.8\% (1.42) \\
\midrule
\multicolumn{4}{l}{\textit{Monte Carlo Q}} \\
Current self-play & \textbf{80.5\% (10.92)} & \textbf{50.2\% (3.65)} & \textbf{32.5\% (0.76)} \\
Checkpoint self-play & 75.4\% (10.35) & 44.1\% (3.18) & 28.8\% (0.12) \\
Smart-only & 77.9\% (10.51) & 46.7\% (3.46) & 29.9\% (0.36) \\
\bottomrule
\end{tabular}%
}
\end{table}

\FloatBarrier

\section{Discussion}
Our results show that direct deep RL can learn useful policies for Big 2, and that algorithm choice matters considerably in the limited-compute setting. In our evaluation, PPO outperforms Monte Carlo Q-approximation, SARSA, and target-network Q-learning under the same simulator, architecture, and training budget. One possible explanation is that value-based methods must estimate noisy, delayed returns for many rare state-action pairs, while PPO can improve the policy from trajectory-level advantage estimates before the value landscape has fully stabilized. In a high-variance, imperfect-information, multiplayer game with non-stationary self-play rewards, this makes value approximation slower to converge and less competitive within the training budget we study.

We also find that controlled stochasticity improves policy learning. PPO without entropy regularization becomes increasingly deterministic, while an intermediate entropy incentive improves performance against Random, Greedy, and Smart opponents. This suggests that imperfect-information card games reward stochastic policies: they encourage exploration during training and give the agent higher success when acting under uncertainty. A prematurely deterministic policy can over-commit to suboptimal action preferences. Our results also show that excessive entropy can make it difficult to exploit learned strategies, meaning that RL approaches must tune the entropy hyperparameter carefully.

The opponent curriculum results show that current-policy self-play is more effective than self-play against previous policy checkpoints or training against the Smart strategy, even when evaluating against the Smart strategy. This result suggests that the best way to exploit a heuristic opponent is not necessarily to train only against that opponent. A fixed deterministic opponent exposes the learner to a narrow slice of the game tree, while current-policy self-play creates an adaptive curriculum whose difficulty tracks the agent's own progress. Checkpoint self-play adds diversity, but under a short training horizon it can dilute the learning signal by spending experience on older opponents the current policy may already beat.

These findings make Big 2 a useful benchmark for studying RL in imperfect-information games. By holding the environment, action representation, architecture, and evaluation protocol fixed, we isolate factors that are often confounded in larger game-playing systems: algorithm choice, policy stochasticity, and opponent distribution. These findings are relevant to real-world multi-agent RL settings, which involve partial observation, delayed rewards, changing opponents, and limited training budgets. Our study is still narrower than the full space of imperfect-information methods: we do not compare against CFR or Deep CFR \citep{zinkevich2007regret,brown2019deepcfr}, nor against search-augmented or opponent-modeling agents. Future work should test whether those methods improve robustness in Big 2, whether value-based methods close the gap with longer training, and whether cross-play among independently trained agents reveals additional strategic weaknesses.

\section*{Impact Statement}
This paper presents work whose goal is to advance reinforcement learning for multiplayer imperfect-information games. We do not deploy the system in real-world decision-making settings; potential societal concerns are limited to the general risks of game-playing AI and strategic agents.

\section*{Acknowledgements}
We thank Vikram Singh for contributing to the game simulator and the training loop and for valuable advice throughout this project.

\bibliography{references}
\bibliographystyle{icml2026}


\newpage
\appendix
\onecolumn
\section{Big 2 Game Details}
\label{app:game_details}
\subsection{Game Definition}
The simulator represents each card as an integer in $\{0,\ldots,51\}$, ordered first by rank and then by suit. Ranks increase in the order 3, 4, 5, 6, 7, 8, 9, 10, J, Q, K, A, 2, and suits increase as diamonds $<$ clubs $<$ hearts $<$ spades. Thus card $0$ is $3\diamondsuit$, the lowest card in the game.

At the beginning of an episode, the deck is uniformly shuffled and dealt evenly, giving each player 13 private cards. The player holding $3\diamondsuit$ acts first and must include that card in the opening play. Players then act clockwise until one player empties their hand. The first player to empty their hand is the winner, and the episode terminates immediately.

\subsection{State, Observations, and Information}
The full simulator state consists of each player's private hand, the current player index, the current active trick, the set of public cards that have already been played, the number of consecutive passes, and the cards played by each player. This full state is not observed by any learning agent. Instead, at a decision point for player $i$, the environment returns an information-state observation containing:
\begin{itemize}
    \item player $i$'s current hand, padded to 13 with pad integers;
    \item a 52-dimensional indicator for the current active trick, if one exists;
    \item a 52-dimensional indicator for all cards seen so far;
    \item the remaining card counts for the other players in clockwise order;
    \item the current number of consecutive passes;
    \item per-opponent 52-dimensional indicators for cards that each opponent has already played.
\end{itemize}
For the standard four-player game, this produces a fixed-length observation vector of dimension
\[
13 + 52 + 52 + 3 + 1 + 3\cdot 52 = 277.
\]
The observation therefore combines the acting player's private hand with public history, but never exposes the unplayed cards in opponents' hands.

\subsection{Actions and Legal Candidate Generation}
At every decision point, the simulator enumerates all candidate combinations in the acting player's current hand. It then filters this set according to the active trick. If there is no active trick, the player has control and may lead any non-pass legal combination. If the active trick is a single, pair, or triple, the player may only play the same type with a higher comparison key. If the active trick is a five-card hand, the player may play a stronger hand in the same category or any legal five-card hand in a higher category. Passing is legal only when there is an active non-pass trick. This produces a variable-length legal action set $\mathcal{A}(o_i)$ for each observation $o_i$.

Big 2's action space is structurally combinatorial in a way that differs from poker-style betting games. In no-limit poker, much of the action-space challenge comes from abstracting over bet sizes. In Big 2, actions are subsets of the player's private hand, and the same card can participate in many incompatible future combinations. Furthermore, each card in the hand is different as both rank and suit matter. This makes fixed action abstraction difficult, because the useful action set depends heavily on the exact cards in the current hand and on the current trick.

This structure makes even local decisions strategically ambiguous. When a player has control and may lead a new trick, the legal set can contain several qualitatively different plans: low singles that probe opponents' responses, pairs or triples that shed duplicated ranks, and five-card hands that may either unlock or destroy future structure. The best lead is therefore not determined only by immediate combination strength. A high card or strong five-card hand can win control now, but spending it may remove the player's only answer to a later threat; conversely, saving it may allow an opponent to seize control. Strong play also depends on hidden-hand inference. Because opponents' hands are observed only through their plays, passes, and remaining card counts, a player must avoid playing tricks that are likely to match an opponent's remaining cards, such as opening a pair or five-card category that lets an opponent shed an otherwise difficult holding.

To quantify this structure, we sampled 10{,}000 complete games using random legal play, yielding 752{,}677 decision points. The visited legal action count is highly state-dependent: many response states are tightly constrained, but the 99th percentile has 19 legal actions and the largest observed decision has 132 legal actions. The branching factor is especially large when a player has control, where the mean legal action count is 8.1 and the 95th percentile is 20.

\subsection{Transition Dynamics}
When a player plays a non-pass combination, the simulator removes those cards from the player's hand, marks them as seen, records them in that player's public played-card history, and sets the active trick to that combination. When a player passes, the consecutive-pass counter increases. If all other players pass after a non-pass play, the active trick is cleared and the last player who played a non-pass combination takes control on the next turn.

\section{Architectural Details}
\label{app:architecture_details}
The state encoder uses a shared card embedding table for all card-valued observation fields. The acting player's hand is represented as a padded list of card ids, encoded with masked self-attention, and pooled over valid cards. The current trick, the set of seen cards, and each opponent's public played-card history are represented as 52-dimensional indicators and projected through the same card embedding table before being passed through small feed-forward encoders. The remaining-card counts for the other players and the current pass count are encoded separately and concatenated with the pooled card features. A projection, layer normalization, and residual feed-forward block produce the final state embedding.

Each legal action is represented by the 80-dimensional candidate feature vector containing the cards involved in the trick and the features of the trick itself, and it is encoded by a two-layer multilayer perceptron. A learned linear projection maps the state embedding into the action-embedding space, and a scaled dot product produces one scalar per legal candidate. In the PPO policy this scalar is a logit; in the Q-network it is $Q(o_i,a)$. The PPO policy additionally applies a multilayer value head to the state embedding to estimate $V(o_i)$.

\section{Training Implementation Details}
\label{app:training_details}
\paragraph{Rollout ownership and seating.}
Each training episode is a full four-player game. A model-controlled seat is a seat whose action is selected by the learned policy or Q-network and whose decision records are used for optimization. In current-policy self-play, the same parameterized policy controls all four seats. Gradients are aggregated from every model-controlled decision point. In fixed-opponent training, a learner seat is sampled for each episode and the remaining seats use the fixed opponent. In checkpoint self-play, non-learner seats are sampled from the current policy or up to 20 saved checkpoints according to the stated mixture, but only learner/model-controlled seats contribute stored training records. Seat assignment is randomized independently of the random deal and the player holding $3\diamondsuit$ still starts the game.

\paragraph{Rewards.}
Rewards are assigned from each model-controlled seat's own perspective. Evaluation uses the unshaped terminal Big 2 score: the winner receives the number of cards left in the other hands, and each loser receives the negative number of cards left in their own hand. Under the score-defined environment reward, intermediate rewards are zero and the terminal score is assigned only when the game ends.

\paragraph{PPO training details.}
PPO uses generalized advantage estimation with $\lambda=0.95$ and $\gamma=0.99$. Advantages are normalized within the update batch. The value loss coefficient is $c_v=0.5$; the entropy coefficient is the $\beta_{\mathrm{ent}}$ reported for each PPO condition, with $\beta_{\mathrm{ent}}=0$ in the no-entropy main comparison. The implementation uses clipped policy ratios with $\epsilon=0.2$, clipped value loss with the same clipping range, and global gradient-norm clipping at $0.5$. Training involves 64 full games per batch, and PPO uses a minibatch size of $256$.

\paragraph{Value-based training details.}
The value-based agents use Adam with learning rate $3\times 10^{-5}$, $\gamma=0.99$, 64 full games per batch, and one optimizer update per collected batch. They do not use a replay buffer; updates are on-policy with respect to the $\epsilon$-greedy behavior policy used to collect that batch. Epsilon decays linearly from $0.5$ to $0$ over training. SARSA and Q-learning use a delayed target network synchronized every 10 batches. The loss is mean squared error on the selected action's Q value. Gradients are clipped to norm $1.0$. Terminal rewards are divided by 13 (the number of cards per hand) to compress into a more natural range for the dot-product scorer.

\section{PPO Entropy Ablation}
\label{app:ppo_entropy_ablation}
\begin{figure}[h]
    \centering
    \includegraphics[width=0.85\textwidth,height=0.55\textheight,keepaspectratio]{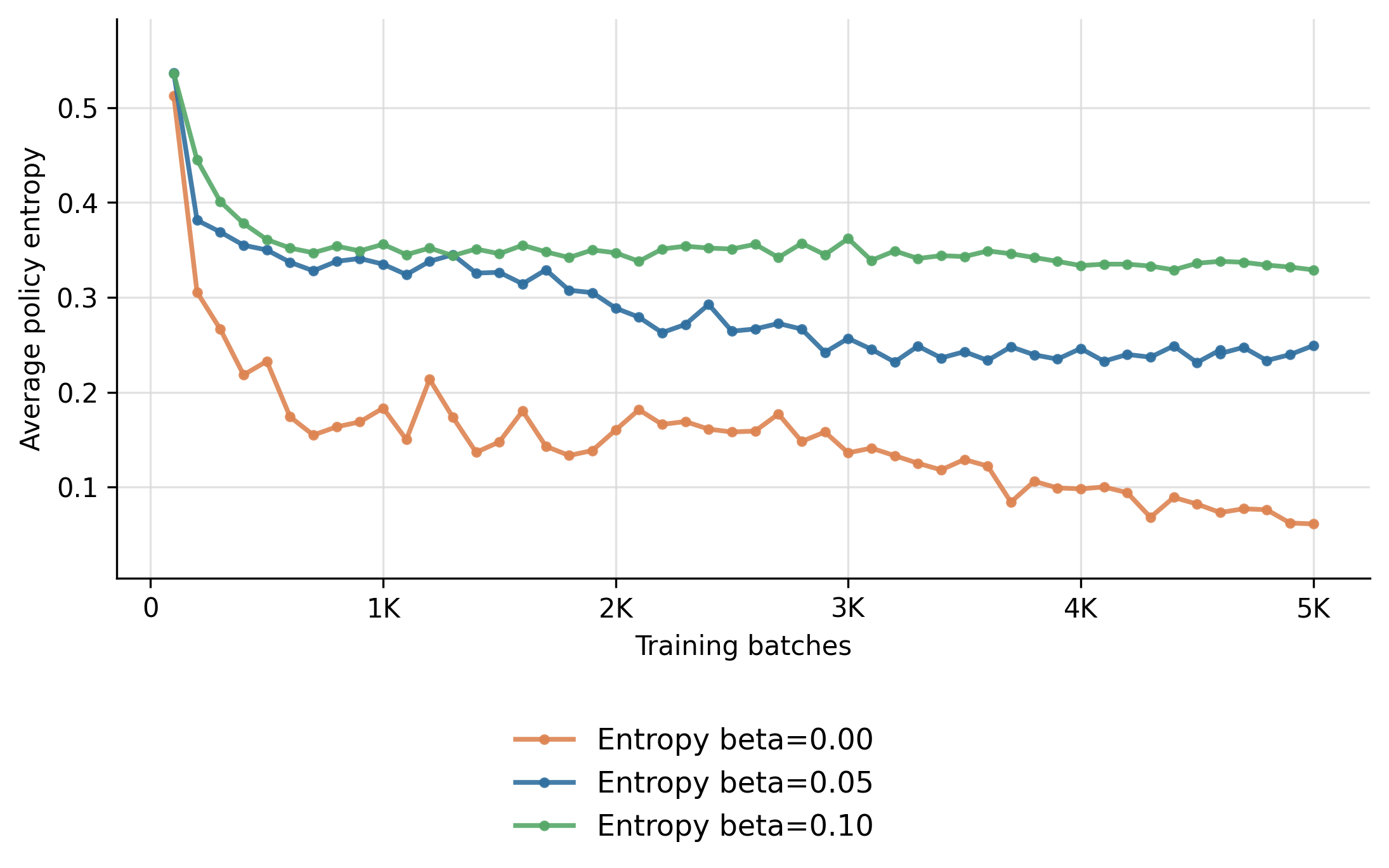}
    \caption{Average policy entropy during PPO current self-play training for different entropy coefficients. Entropy is sampled every 100 training batches. Larger entropy coefficients keep the policy more stochastic over the course of training, while the run with no entropy bonus becomes increasingly deterministic.}
    \label{fig:ppo_entropy_ablation}
\end{figure}

\section{Heuristic Baselines}
\subsection{Greedy Heuristic Baseline}
The greedy baseline is a deterministic rule-based policy used as a simple non-learning opponent. If the player is forced to take the only legal action, the policy returns it. Otherwise, it excludes \textsc{Pass} and chooses the minimum non-pass candidate under the simulator's combination ordering. This ordering sorts first by combination type and then by the combination comparison key, so the policy plays the weakest legal non-pass action available rather than preserving hand structure or reasoning about future tricks.

\begin{algorithm}[h]
\caption{Greedy heuristic action selection}
\label{alg:greedy_strategy}
\begin{algorithmic}[1]
\REQUIRE legal candidates $\mathcal{A}$
\IF{$|\mathcal{A}| \le 1$}
    \STATE \textbf{return} the only legal action in $\mathcal{A}$
\ENDIF
\STATE $\mathcal{B} \gets \{a \in \mathcal{A}: a \neq \textsc{Pass}\}$
\STATE \textbf{return} $\min_{a\in\mathcal{B}} a$ under the simulator's combination ordering
\end{algorithmic}
\end{algorithm}

\subsection{Smart Heuristic Strategy}
The Smart strategy is a deterministic rule-based policy used as a stronger non-learning opponent. It scores each non-pass legal action and chooses the minimum-scoring action, with lower scores corresponding to more desirable plays. The heuristic favors immediate wins, shedding more cards, preserving future combinations, avoiding early use of 2s, and avoiding low orphan cards. Passing is considered only in narrow cases: when the best early-game response would spend multiple 2s, or when the current trick is a four-of-a-kind or straight flush.

\begin{algorithm}[h]
\caption{Smart heuristic action selection}
\label{alg:smart_strategy}
\begin{algorithmic}[1]
\REQUIRE legal candidates $\mathcal{A}$, current hand $H$, active trick $T$
\STATE $\mathcal{B} \gets \{a \in \mathcal{A}: a \neq \textsc{Pass}\}$
\IF{$\mathcal{B}=\emptyset$ or $|\mathcal{A}|=1$}
    \STATE \textbf{return} the only legal action
\ENDIF
\FORALL{$a \in \mathcal{B}$}
    \IF{$|a|=|H|$}
        \STATE $s(a) \gets -1000$ \COMMENT{win immediately}
    \ELSE
        \STATE $s(a) \gets 0.8\sum_{c\in a}\mathrm{rank}(c)$
        \IF{phase$(H)$ is early} \STATE $s(a) \gets s(a)+10\cdot\#\{2\text{s in }a\}$ \ENDIF
        \IF{phase$(H)$ is mid} \STATE $s(a) \gets s(a)+5\cdot\#\{2\text{s in }a\}$ \ENDIF
        \STATE $s(a) \gets s(a)+\mathrm{BreakPenalty}(a,H)$
        \STATE $s(a) \gets s(a)+6\cdot\mathrm{LowOrphans}(H\setminus a)$
        \STATE $s(a) \gets s(a)-4|a|$
        \IF{phase$(H)$ is late} \STATE $s(a) \gets s(a)-10$ \ENDIF
        \IF{$T$ is very strong and phase$(H)$ is late} \STATE $s(a) \gets s(a)-10$ \ENDIF
        \IF{$T$ is four-of-a-kind or straight flush} \STATE $s(a) \gets s(a)+25$ \ENDIF
    \ENDIF
\ENDFOR
\STATE $a^\star \gets \arg\min_{a\in\mathcal{B}} s(a)$
\IF{\textsc{Pass} $\in \mathcal{A}$ and phase$(H)$ is early and $a^\star$ uses at least two 2s and $s(a^\star)>30$}
    \STATE \textbf{return} \textsc{Pass}
\ENDIF
\IF{\textsc{Pass} $\in \mathcal{A}$ and $T$ is four-of-a-kind or straight flush}
    \STATE \textbf{return} \textsc{Pass}
\ENDIF
\STATE \textbf{return} $a^\star$
\end{algorithmic}
\end{algorithm}

\paragraph{Break penalty.}
\textsc{BreakPenalty} returns $8$ in the early game or $4$ in the mid game when an action breaks a remaining pair or triple. When an action breaks a potential five-card structure, the penalty is $20$ in the early game, $8$ in the mid game, and $4$ in the late game. The implementation checks four-of-a-kind, full-house components, flushes with at least five cards of a suit, and five-rank straight windows excluding rank 2.

\paragraph{Game phases.}
The heuristic defines early, mid, and late game by the acting player's hand size: early if $|H|>10$, mid if $6 \le |H| \le 10$, and late if $|H|\le 5$.

\end{document}